\documentclass[10pt,letterpaper]{article}
\usepackage[top=0.85in,left=2.75in,footskip=0.75in]{geometry}

\usepackage{amsmath,amssymb}
\usepackage{adjustbox}

\usepackage{changepage}

\usepackage{textcomp,marvosym}

\usepackage{cite}

\usepackage{nameref,hyperref}
\usepackage{booktabs}
\usepackage{algorithmic}

\usepackage[right]{lineno}

\usepackage[nopatch=eqnum]{microtype}
\DisableLigatures[f]{encoding = *, family = * }

\usepackage[table]{xcolor}
\usepackage{algorithm}

\usepackage{array}
\newcolumntype{+}{!{\vrule width 2pt}}

\newlength\savedwidth

\usepackage[aboveskip=1pt,labelfont=bf,labelsep=period,justification=raggedright,singlelinecheck=off]{caption}

\makeatletter
\renewcommand{\@biblabel}[1]{\quad#1.}
\makeatother

\usepackage{longtable}
\usepackage{changepage}

\usepackage{subcaption}

\usepackage{lastpage,fancyhdr,graphicx}
\usepackage{epstopdf}
\fancyheadoffset[L]{2.25in}
\fancyfootoffset[L]{2.25in}
\begin{document}
\vspace*{0.2in}

\begin{flushleft}
{\Large
\textbf\newline{Using Pre-training and Interaction Modeling for ancestry-specific disease prediction using multiomics data from the UK Biobank} 
}
\newline
\\
Thomas Le Menestrel\textsuperscript{1*},
Erin Craig\textsuperscript{3},
Robert Tibshirani\textsuperscript{2,3},
Trevor Hastie\textsuperscript{2},
Manuel Rivas\textsuperscript{3*}
\\
\bigskip
\textbf{1}: Institute for Computational and Mathematical Engineering (ICME), School of Engineering, Stanford University, Stanford, CA, United States of America 
\textbf{2} Department of Statistics, Stanford University, Stanford, CA, United States of America 
\textbf{3} Department of Biomedical Data Science, Stanford University, Stanford, CA, United States of America \\
\bigskip

* tlmenest@stanford.edu (TLM); mrivas@stanford.edu (MAR)

\end{flushleft}
\section*{Abstract}

Recent genome-wide association studies (GWAS) have uncovered the genetic basis of complex traits, but show an under-representation of non-European descent individuals, underscoring a critical gap in genetic research. Here, we assess whether we can improve disease prediction across diverse ancestries using multiomic data. We evaluate the performance of Group-LASSO INTERaction-NET (glinternet) and pretrained lasso in disease prediction focusing on diverse ancestries in the UK Biobank. Models were trained on data from White British and other ancestries and validated across a cohort of over 96,000 individuals for 8 diseases. Out of 96 models trained, we report 16 with statistically significant incremental predictive performance in terms of ROC-AUC scores (\(p\text{-value} < 0.05\)), found for diabetes, arthritis, gall stones, cystitis, asthma and osteoarthritis. For the interaction and pretrained models that outperformed the baseline, the PRS score was the primary driver behind prediction. Our findings indicate that both interaction terms and pre-training can enhance prediction accuracy but for a limited set of diseases and moderate improvements in accuracy. Our code is available at (\url{https://github.com/rivas-lab/AncestryOmicsUKB}).

\section*{Author summary}
Glinternet and pretrained lasso are statistical methods that excel in scenarios where data for one population is limited, by leveraging common patterns from a larger population through interaction modelling and pre-training. This capability is crucial for the UK Biobank where 75\% of the data pertains to White British individuals while certain ancestries make up less than 1\% of the database. Despite the widespread use of LASSO and Cox proportional hazard models \cite{cox1972regression, Tibshirani1996}, strategies that consider interactions and pre-training remain unexplored. In this study, we trained glinternet and pretrained lasso models on mixed datasets of White British and other ancestries including South Asian, African, non-British European and Admixed across 8 diseases, reporting 16 models with significant improvements (\(p\text{-value} < 0.05\)) out of 96. 

\section*{Introduction}

Genome-wide association studies (GWAS) have significantly advanced our understanding of the genetic basis of complex traits and diseases \cite{cano2020gwas, gondro2013genome}. Yet, these studies have predominantly focused on individuals of European descent, leading to under-representation of non-European ancestry in genomic databases \cite{Popejoy2016, peterson2019genome}, despite ancestry being a crucial factor in disease susceptibility \cite{kittles2003race}. This is also the case for the UK Biobank, which contains approximately 75\% data from White British individuals and less than 10\% from diverse ancestries such as Related (9.9\%), Admixed (6.4\%), Non-British European (5.6\%), South Asian (1.8\%), African (1.5\%) and East Asian (0.4\%) \cite{Bycroft2018}. 

Efforts to address the challenge of ancestry-specific disease prediction have varied. Polygenic risk scores (PRS) were trained for disease prediction using UK Biobank data \cite{Tanigawa2022, SinnottArmstrong2021}; however, their efficacy diminishes when applied to non-European individuals. PCA approaches have also been developed to create ancestry-specific PRS scores for disease prediction \cite{Kullo2020}. Alternative strategies employing LASSO variants and advanced machine learning techniques like AutoEncoders have been attempted \cite{Lehmann2023, Gyawali2023}. Yet, these methods struggle with non-linearity and lack the interpretability of simpler models. LASSO and Cox proportional hazard models, as well as penalized variants, have been extensively used for this problem but under-perform when dealing with non-linearity as well as because of the proportional hazard assumption failing in practice \cite{goeman2010l1, li2022fast, bi2020fast, Widen2021}. Finally, how multiomics data beyond genetics improve disease risk prediction in non-European populations is unclear and most studies have focused on single data entity as well as single diseases \cite{CarrascoZanini2023, Gadd2023, Barrett}.

Addressing these limitations, our study uses pre-training and interaction modeling through glinternet and pretrained lasso across multiple ancestries and individual ancestries jointly, and facilitate the application of patterns identified in White British data to other ancestries. Interaction modeling allows to adjust coefficients of certain variables based on specific covariates, often age, sex or PRS score for a given disease. This allows to capture non-linear relationships between features, for instance, the effect of a genetic variant which might be different depending on the ancestry of a patient or environmental factors. On the other hand, pre-training on a large dataset with multiple ancestries is particularly beneficial when underlying genetic patterns may be complex and not easily discernible from smaller datasets. After the initial broad training, the model is fine-tuned on a smaller, more specific dataset. This is crucial when dealing with PRS scores for diverse ancestries because it allows the model to adapt to the nuances of specific genetic backgrounds that might not be well-represented in the larger dataset.

We compiled clinical data from over 96,000 individuals spanning 8 diseases with demographic, metabolomic, genetic and biomarker data, thus having a dataset with multiomic data and multiple diseases, addressing the issues seen in previous studies. This approach improves prediction accuracy across a diverse range of ancestries and maintains the level of interpretability achieved in previous studies. 

\section*{Ethics statement}

This research has been conducted using the UK Biobank Resource under Application Number 24983, “Generating effective therapeutic hypotheses from genomic and hospital linkage data” (http://www.ukbiobank.ac.uk/wp-content/uploads/2017/06/24983-Dr-Manuel-Rivas.pdf). Based on the information provided in Protocol 44532, the Stanford IRB has determined that the research does not involve human subjects as defined in 45 CFR 46.102(f) or 21 CFR 50.3 (g). All participants of the UK Biobank provided written informed consent (information available at https://www.ukbiobank.ac.uk/2018/02/gdpr/).

\newpage

\section*{Results}

\subsection*{Results for African ancestry}

For African ancestry, we find 6 statistically significant models (\(p\text{-value} < 0.05\)), which are all glinternet models (cystitis, gall stones, diabetes and arthritis). In terms of average ROC-AUC score across all diseases, pretrained lasso shows suboptimal performance and is similar or worse than the baseline models depending on the subset used for training, while glinternet slighty outperforms both the logistic regression and pretrained lasso models (Tables 2 and 3).

The glinternet model shows significant improvements over the baseline for several diseases such as cystitis ($\Delta$ ROC-AUC of 0.039, 7.1\% increase), gall stones ($\Delta$ ROC-AUC of 0.054, 11.2 \% increase), diabetes ($\Delta$ ROC-AUC of 0.022, 2.5 \% increase) and arthritis ($\Delta$ ROC-AUC of 0.073, 11.4 \% increase). The model is consistent across all diseases and is only slighty outperformed for myocardial infarction and osteoarthritis by a small margin (less than 0.01 $\Delta$ ROC-AUC). The glinternet models trained on the Mix datasets have an average ROC-AUC 0.028 higher than the baselines (4.1 \% increase).

We find that L1-penalized Logistic Regression keeps 147 non-zero coefficients when trained on the Mix dataset for diabetes, while Pretrained LASSO keeps only 19 of them and outperforms it by a $\Delta$ ROC-AUC of 0.02. For Glinternet for the same disease, we report a $\Delta$ ROC-AUC of 0.022 and find 60 interaction terms (27 with age, 25 with the PRS score for diabetes and 8 with sex).

Glinternet and logistic regression provide the best accuracy for predicting certain diseases in individuals of African ancestry, specifically for cystitis, gall stones, diabetes and arthritis, as shown by the ROC-AUC scores and the p-values of the ROC tests. On the contrary, pretrained lasso does not show statistically significant improvements for the ROC tests and has same or lower average ROC-AUC scores than both baselines (Tables 1 and 2).

\begin{table}[ht]
\begin{adjustwidth}{-1.85in}{0in}
\centering
\caption{ROC-AUC scores for logistic regression, pretrained lasso and glinternet (African ancestry).}
\begin{tabular}{llcccccccccccc}
\hline
Method & Data & CRF & CYS & GL & DIA & MI & OST & AST & ART & Avg. & Inc. \\
\hline
Logistic regression & AF & 0.718 & 0.5 & 0.5 & 0.844 & 0.5 & 0.797 & 0.622 & 0.718 & 0.65 & 0 \\
Logistic regression & Mix & 0.71 & 0.549 & 0.474 & 0.839 & 0.753 & 0.847 & 0.629 & 0.641 & 0.68 & 0.03 \\
Logistic regression & All & 0.705 & 0.549 & 0.471 & 0.854 & \textbf{0.755} & \textbf{0.848} & 0.643 & 0.683 & 0.688 & 0.038 \\
\hline
Glinternet & Mix & 0.706 & \textbf{0.588} & \textbf{0.527} & 0.86 & 0.754 & 0.84 & \textbf{0.675} & 0.714 & \textbf{0.708} & \textbf{0.058} \\
Glinternet & All & 0.718 & 0.526 & 0.501 & \textbf{0.872} & 0.745 & 0.839 & 0.667 & \textbf{0.741} & 0.701 & 0.051 \\
\hline
Pretrained lasso& Mix & 0.689 & 0.568 & 0.478 & 0.858 & 0.63 & 0.788 & 0.639 & 0.697 & 0.669 & 0.018 \\
Pretrained lasso& All & \textbf{0.725} & 0.567 & 0.477 & 0.865 & 0.646 & 0.838 & 0.621 & 0.701 & 0.68 & 0.03 \\
\hline
\end{tabular}
\end{adjustwidth}
\label{tab:results_combined}
\end{table}

\begin{table}[ht]
\begin{adjustwidth}{-1.25in}{0in}
\centering
\caption{P-values and $\Delta$ ROC-AUC for pretrained lasso and glinternet against baselines (African ancestry).}
\begin{tabular}{lllcccccccccc}
\hline
Method & Metric & Data & CRF & CYS & GL & DIA & MI & OST & AST & ART \\
\hline
Glinternet & P-value & Mix & 0.572 & \textbf{0.016} & \textbf{0.033} & \textbf{0.002} & 0.470 & 0.704 & 0.073 & \textbf{0.018} \\
Glinternet & $\Delta$ AUC & Mix & -0.004 & \textbf{0.039} & \textbf{0.054} & \textbf{0.022} & 0.001 & -0.008 & 0.046 & \textbf{0.073} \\
\hline
Glinternet & P-value & All & 0.250 & 0.653 & 0.079 & \textbf{0.025} & 0.729 & 0.728 & 0.166 & \textbf{0.012} \\
Glinternet & $\Delta$ AUC & All & 0.013 & -0.023 & 0.030 & \textbf{0.018} & -0.010 & -0.009 & 0.024 & \textbf{0.058} \\
\hline
Pretrained lasso & P-value & Mix & 0.655 & 0.463 & 0.460 & 0.073 & 0.844 & 0.920 & 0.418 & 0.101 \\
Pretrained lasso& $\Delta$ AUC & Mix & -0.021 & 0.019 & 0.005 & 0.02 & -0.123 & -0.059 & 0.01 & 0.056 \\
\hline
Pretrained lasso& P-value & All & 0.274 & 0.221 & 0.440 & 0.138 & 0.948 & 0.735 & 0.665 & 0.176 \\
Pretrained lasso & $\Delta$ AUC & All & 0.02 & 0.018 & 0.006 & 0.01 & -0.109 & -0.009 & -0.022 & 0.018 \\
\hline
\end{tabular}
\end{adjustwidth}
\label{tab:results_combined}
\end{table}

\newpage
\subsection*{Results for South Asian ancestry}

For South Asian ancestry, we find 8 statistically significant models (\(p\text{-value} < 0.05\)): five glinternet models (osteoarthritis, arthritis, asthma) and 3 pretrained lasso models (asthma and arthritis). Both glinternet and pretrained lasso show improved performance in terms of $\Delta$ ROC-AUC compared to the baselines trained on the ancestry-only and Mix datasets and produce the top performing ROC-AUC scores for all diseases except for cystitis and diabetes (Tables 4 and 5).

Both models perform particularly well for arthritis, with each combination of model and dataset giving a statistically significant p-value for the ROC tests. $\Delta$ ROC-AUCs for those disease range from 0.062 (9.6 \% increase) to 0.095 (15 \% increase)

We find that L1-penalized Logistic Regression keeps 73 non-zero coefficients for arthritis, while pretrained lasso keeps only 2 of them, the PRS score and sex. It outperforms it by a $\Delta$ ROC-AUC of 0.095. For glinternet for the same disease, we report a $\Delta$ ROC-AUC of 0.091 and find 60 interaction terms (27 with age, 25 with the PRS score for diabetes and 8 with sex).

The pretrained lasso models trained on the Mix dataset have similar predictive ability compared to the baselines, while the best performing model in terms of average ROC-AUC score is pretrained lasso trained on the All dataset. The glinternet models perform similarly well in terms of average ROC-AUC, trained on either the Mix or All datasets.  Asthma and osteoarthritis both have two significant p-values. Others combinations for different diseases show superior or on par performance with the baselines in terms of average ROC-AUC scores and p-values close to the threshold for statistical significance.

\begin{table}[ht]
\centering
\begin{adjustwidth}{-1.70in}{0in}
\caption{ROC-AUC scores for logistic regression, pretrained lasso and glinternet (South Asian ancestry).}
\begin{tabular}{llcccccccccccc}
\hline
Method & Data & CRF & CYS & GL & DIA & MI & OST & AST & ART & Avg. & Inc. \\
\hline
Logistic regression & SA & 0.748 & 0.5 & 0.5 & 0.782 & 0.675 & 0.706 & 0.5 & 0.706 & 0.639 & 0 \\
Logistic regression & Mix & 0.834 & \textbf{0.663} & 0.629 & 0.802 & 0.703 & 0.688 & 0.624 & 0.612 & 0.694 & 0.055 \\
Logistic regression & All & 0.83 & 0.555 & 0.635 & \textbf{0.815} & 0.714 & 0.691 & 0.639 & 0.646 & 0.691 & 0.051 \\
\hline
Glinternet & Mix & 0.834 & 0.631 & \textbf{0.659} & 0.814 & 0.718 & 0.736 & \textbf{0.652} & 0.703 & 0.718 & 0.079 \\
Glinternet & All & \textbf{0.84} & 0.629 & 0.648 & 0.813 & 0.725 & \textbf{0.739} & 0.649 & 0.708 & 0.719 & 0.079 \\
\hline
Pretrained lasso & Mix & 0.829 & 0.63 & 0.551 & 0.807 & 0.746 & 0.714 & 0.642 & 0.706 & 0.703 & 0.064 \\
Pretrained lasso & All & 0.84 & 0.634 & 0.648 & 0.811 & \textbf{0.75} & 0.713 & 0.645 & \textbf{0.726} & \textbf{0.721} & \textbf{0.081} \\
\hline
\end{tabular}
\end{adjustwidth}
\label{tab:results_combined}
\end{table}

\begin{table}[ht]
\begin{adjustwidth}{-1.2in}{0in}
\centering
\caption{P-values and $\Delta$ ROC-AUC for pretrained lasso and glinternet against logistic regression (South Asian ancestry).}
\begin{tabular}{lllcccccccc}
\hline
Method & Metric & Data & CRF & CYS & GL & DIA & MI & OST & AST & ART \\
\hline
Glinternet & P-value & Mix         & 0.516  & 0.702  & 0.118 & 0.097 & 0.142 & \textbf{0.005} &  \textbf{0.070} & \textbf{0.007} \\
Glinternet & $\Delta$ AUC  & Mix   & -0.001 & -0.032 & 0.03 & 0.012 & 0.015 & \textbf{0.048} & \textbf{0.028} & \textbf{0.091} \\
\hline
Glinternet & P-value & All & 0.157 & 0.074 & 0.221 & 0.607 & 0.210 & \textbf{0.015} & 0.273 & \textbf{0.020} \\
Glinternet & $\Delta$ AUC & All & 0.01 & 0.075 & 0.013 & -0.003 & 0.01 & \textbf{0.047} & 0.01 & \textbf{0.062} \\
\hline
Pretrained lasso & P-value & Mix & 0.747 & 0.708 & 0.772 & 0.334 & 0.113 & 0.226 & \textbf{0.039} & \textbf{0.023} \\
Pretrained lasso & $\Delta$ AUC & Mix & -0.005 & -0.033 & -0.078 & 0.004 & 0.043 & 0.026 & \textbf{0.018} & \textbf{0.095} \\ 
\hline
Pretrained lasso & P-value & All & 0.200 & 0.064 & 0.391 & 0.746 & 0.129 & 0.168 & 0.187 & \textbf{0.016} \\
Pretrained lasso & $\Delta$ AUC & All & 0.01 & 0.079 & 0.013 & -0.005 & 0.036 & 0.022 & 0.006 & \textbf{0.08} \\
\hline
\end{tabular}
\end{adjustwidth}
\label{tab:results_combined}
\end{table}

\newpage

\subsection*{Results for Admixed ancestry}

For Admixed ancestry, we find only 2 statistically significant models (\(p\text{-value} < 0.05\)), both of which are pretrained lasso models for gall stones. We find that the Admixed glinternet and pretrained lasso models both perform better than the baselines but do not show statistically significant improvements (Tables 6 and 7). Pretrained lasso outperforms both baselines by a $\Delta$ ROC-AUC ranging from 0.036 (4.8 \% increase) for both datasets. Both glinterned and pretrained lasso are the top performing models in terms of average ROC-AUC except for myocardial infarction and chronic renal failure.

We find that L1-penalized Logistic Regression keeps 94 non-zero coefficients for ostheoarthritis, while pretrained lasso keeps only 4 of them, the PRS score for ostheoarthritis and Global PCs 1, 2 and 4. It outperforms it by a $\Delta$ ROC-AUC of 0.018. For glinternet for the same disease we report a $\Delta$ ROC-AUC of 0.054 and find 46 interaction terms (1 with ancestry, 10 with age, 10 with the PRS score for ostheoarthritis and 10 with sex).

\begin{table}[ht]
\centering
\begin{adjustwidth}{-1.75in}{0in}
\caption{ROC-AUC scores for logistic regression, pretrained lasso and glinternet (Admixed Analysis).}
\begin{tabular}{llcccccccccccc}
\hline
Method & Data & CRF & CYS & GL & DIA & MI & OST & AST & ART & Avg. & Inc. \\
\hline
Logistic regression & AD & 0.883 & 0.5 & 0.664 & 0.902 & \textbf{0.816} & 0.681 & 0.641 & 0.5 & 0.698 & 0 \\
Logistic regression & Mix & \textbf{0.91} & 0.666 & 0.668 & 0.882 & 0.774 & 0.682 & 0.638 & 0.651 & 0.734 & 0.036 \\
Logistic regression & All & 0.909 & 0.666 & 0.675 & 0.892 & 0.777 & 0.7 & 0.645 & 0.672 & 0.742 & 0.044 \\
\hline
Glinternet & Mix & 0.906 & \textbf{0.692} & 0.711 & 0.885 & 0.793 & 0.727 & 0.663 & 0.653 & \textbf{0.754} & 0.055 \\
Glinternet & All & 0.892 & 0.681 & 0.667 & 0.884 & 0.795 & \textbf{0.754} & \textbf{0.673} & 0.693 & 0.755 & 0.056 \\
\hline
Pretrained lasso & Mix & 0.905 & 0.685 & 0.84 & \textbf{0.948} & 0.752 & 0.7 & 0.656 & 0.67 & 0.77 & 0.071 \\
Pretrained lasso & All & 0.906 & 0.65 & \textbf{0.866} & 0.947 & 0.757 & 0.719 & 0.666 & \textbf{0.706} & 0.777 & \textbf{0.079} \\
\hline
\end{tabular}
\label{tab:results_ot}
\end{adjustwidth}
\end{table}

\begin{table}[ht]
\begin{adjustwidth}{-1.2in}{0in}
\centering
\caption{P-values and $\Delta$ ROC-AUC for pretrained lasso and glinternet against logistic regression (Admixed ancestry).}
\begin{tabular}{lllcccccccc}
\hline
Method & Metric & Data & CRF & CYS & GL & DIA & MI & OST & AST & ART \\
\hline
Glinternet & P-value & Mix & 0.673 & 0.080 & 0.167 & 0.421 & 0.356 & 0.083 & 0.684 & 0.299 \\
Glinternet & $\Delta$ AUC & Mix &-0.003 & 0.026 & 0.043 & 0.002 & 0.019 & 0.045 & 0.024 & 0.002 \\
\hline
Glinternet & P-value & All & 0.877 & 0.176 & 0.174 & 0.656 & 0.381 & 0.064 & 0.670 & 0.285 \\
Glinternet & $\Delta$ AUC & All & -0.017 & 0.015 & -0.008 & -0.007 & 0.018 & 0.054 & 0.028 & 0.021 \\
\hline
Pretrained lasso & P-value & Mix & 0.630 & 0.175 & \textbf{0.001} & 0.092 & 0.570 & 0.320 & 0.787 & 0.269 \\
Pretrained lasso & $\Delta$ AUC & Mix & -0.004 & 0.02 & \textbf{0.172} & 0.066 & -0.023 & 0.018 & 0.018 & 0.018 \\
\hline
Pretrained lasso & P-value & All & 0.764 & 0.761 & \textbf{0.001} & 0.282 & 0.656 & 0.316 & 0.748 & 0.520 \\
Pretrained lasso & $\Delta$ AUC & All & -0.003 & -0.016 & \textbf{0.19} & 0.055 & -0.02 & 0.018 & 0.02 & 0.034 \\
\hline
\end{tabular}
\end{adjustwidth}
\label{tab:results_combined}
\end{table}

\subsection*{Runtimes}

Glinternet models took on average 1h48 minutes to train on the Mix datasets and 2h32 minutes on the All datasets. The L1-penalized logistic regression ones trained using glmnet took 16 minutes to train on the Mix datasets and 23 minutes on the All datasets. Finally, the pretrained lasso models took 41 minutes to train on the Mix datasets and 1h27 minutes on the All datasets. 

\newpage

\subsection*{Results of glinternet for osteoarthritis for South Asian ancestry}

We focus on the results of the glinternet model for osteoarthritis trained with data from White British and South Asian ancestry to understand what is driving the prediction behind our models. 

On Figure 1, the cross-validation error starts to significantly decrease while lowering the level of regularization. Main effects are stable across different levels of regularization, suggesting that there are key predictors with strong associations with the outcome. As regularization is relaxed, more interaction terms are included in the model. There is a significant jump in the number of interaction terms between index 35 and 50 (from 30 to 150 interaction variables), suggesting overfitting past a certain decrease in regularization strength.

As seen on Figure 2, the PRS score is the main covariate from which interaction terms are found. The largest interaction term found is between the PRS score for osteoarthritis and the PC4. It has a coefficient of 26.07 and is three times larger than the second largest one (variables are standardized by glinternet when fitting). In a previous study, it was found that the 4th genetic PC of the UK Biobank helped to distinguish between the different South Asian groups (Indian, Pakistani, Bangladeshi and other South Asian and mixed backgrounds) \cite{2015GenotypingAQ}. This could mean different subgroups within the South Asian ancestry have varying predispositions to osteoarthritis.  

\begin{figure}[htbp]
\centering
\begin{adjustwidth}{-2.35in}{0in}
\begin{minipage}{1\textwidth}
\includegraphics[width=1.5\textwidth]{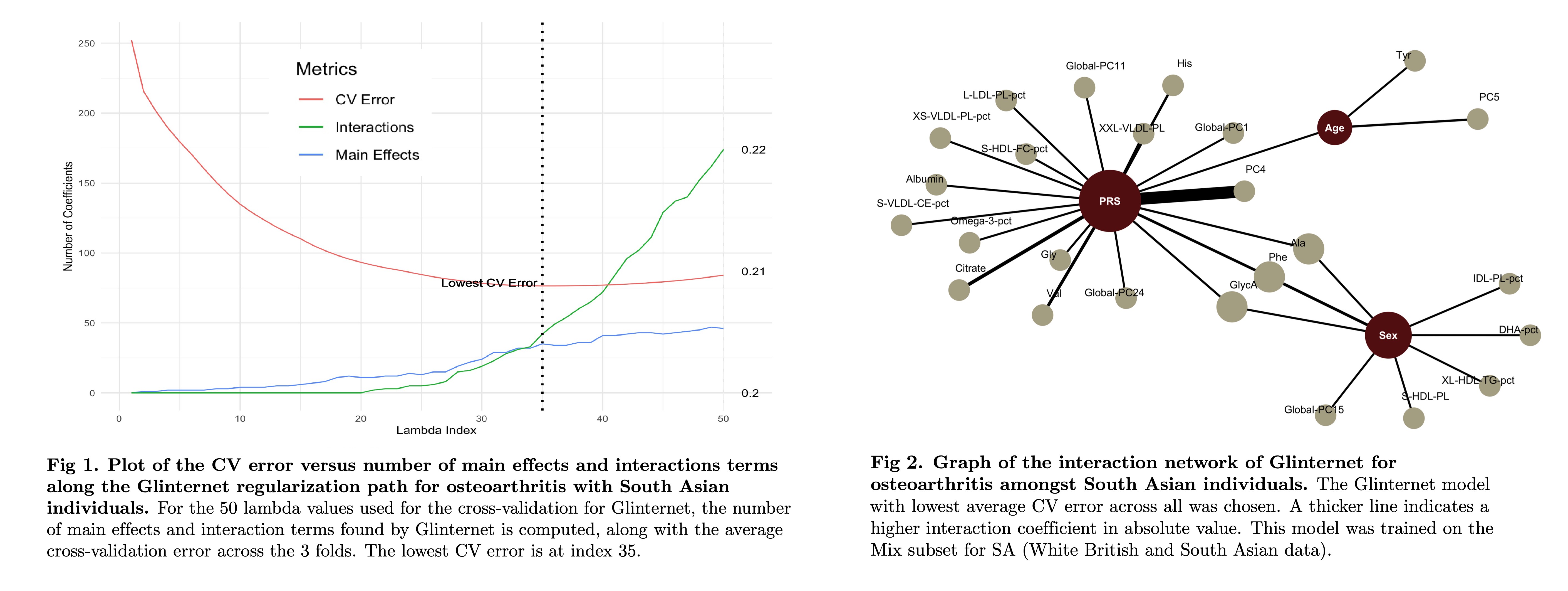}
\label{fig:S1_Fig}
\end{minipage}
\end{adjustwidth}
\end{figure}

\section*{Discussion}

For 96 models trained, 16 had statistically significant incremental predictive performance in terms of ROC-AUC scores (\(p\text{-value} < 0.05\)), specifically for diabetes, arthritis, gall stones, cystitis, asthma and osteoarthritis. This could indicate a strong genetic and ancestral component for the prediction of those diseases. Our findings indicate that both interaction terms and pre-training can enhance prediction accuracy but for a limited set of diseases and moderate improvements in accuracy. 

Pretrained lasso models were much more sparse than their L1-penalized logistic regression counterparts, making them easier to interpret and thus a superior candidate for prediction of certain diseases as they showed superior performance on average for South Asian and Admixed ancestries. They also performed better in terms of average ROC-AUC scores and p-values when trained on the All datasets versus the Mix ones.

We saw minor or even no significant improvements with the pretrained lasso models for African ancestry. This could indicate that people from the African ancestry exhibit distinct statistical characteristics compared to other subgroups, suggesting significant variations in the dataset and making it more difficult to generalize patterns from other ancestries. 

Regarding the analysis of glinternet for osteoarthritis for South Asian ancestry, adding more interaction terms between the PRS score and other variables helped reach optimal performance in terms of cross-validation score, as main effects were still stable when relaxing the level of regularization. The main interaction coefficient driving the prediction was between the PRS score and the 4th PC, which was used in previous studies to distinguish betweeen different ethnicities within the South Asian ancestry.

Finally, both glinternet and pretrained lasso models took a significant amount of time to train compared to the L1-penalized logistic regression baselines. This can be explained by several factors. A pretrained lasso model is made of one \textit{glmnet} model trained on all available ancestries and \textit{glmnet} sub-models for each ancestry used. This can make training computationally expensive, as each new ancestry added requires a separate sub-model as well as increases the size of the training set for the overall model. For glinternet, as it evaluates pairwise variable interactions for the Group-LASSO model, the complexity scales up significantly when evaluating a large number of candidate interaction terms.

\section*{Materials and methods}

\setcounter{figure}{2} 

\begin{figure}[ht]
\centering
\begin{adjustwidth}{-1.05in}{0in}
\begin{minipage}{1\textwidth}
\includegraphics[width=1.2\textwidth]{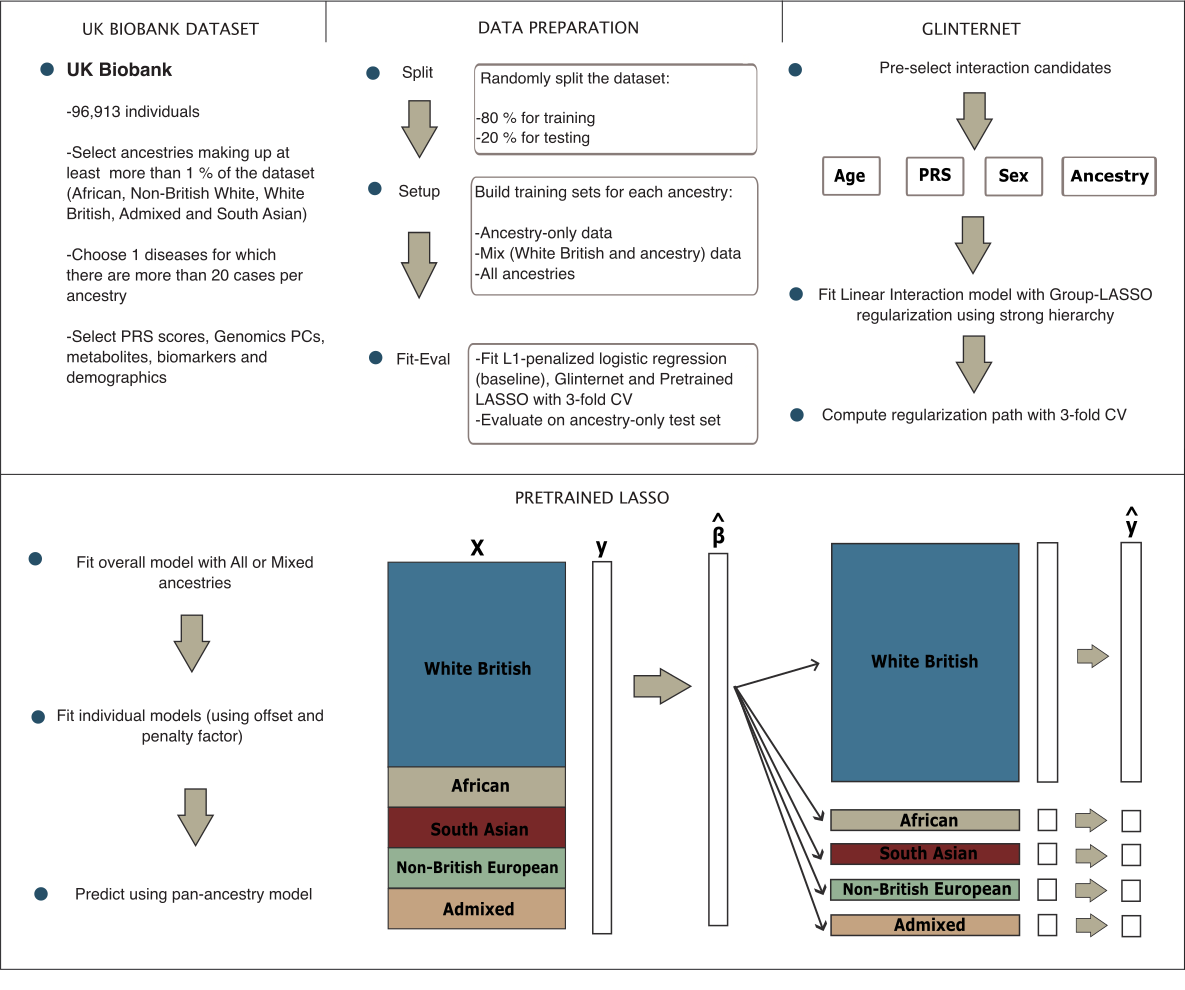}
\caption{{{\bf Diagram of the methodology followed for this project.}}}
\label{fig:interactions}
\end{minipage}
\end{adjustwidth}
\end{figure}

\subsection*{L1-penalized Logistic Regression}

The Lasso regression in the context of a binomial response variable can be expressed as an optimization problem. The objective is to minimize the cost function that comprises the log-likelihood of a binomial distribution along with an \(L_1\) penalty term. The mathematical formulation is given by:

\begin{equation}
    \min_{\beta} \left( -\sum_{i=1}^n \left[ y_i \log(p_i) + (1 - y_i) \log(1 - p_i) \right] + \lambda \sum_{j=1}^p |\beta_j| \right)
\end{equation}

where \(\beta\) represents the coefficients of the model, \(n\) is the number of observations, \(y_i\) is the response variable which can take values 0 or 1, \(p_i\) is the probability of \(y_i\) being 1, often modeled as \(p_i = \frac{1}{1 + e^{-\beta^T x_i}}\), where \(x_i\) is the vector of explanatory variables for the \(i\)-th observation and \(\lambda\) is the regularization parameter, controlling the strength of the \(L_1\) penalty term.

\subsection*{Glinternet}

Glinternet  is a comprehensive framework for fitting first-order interaction models that respect a strong hierarchy, meaning an interaction effect between two variables is considered in the model only if both main effects of the interacting variables are also included. This approach is particularly effective in the context of high-dimensional data, where interactions between variables can provide significant insights but also pose challenges in terms of model complexity and interpretability. \\The fundamental model for a binary response \( Y \) is given by:

\begin{equation}
\text{logit}(P(Y = 1|X)) = \mu + \sum_{i=1}^{p} \left( X_i \theta_i + \sum_{j>i} X_{i:j} \theta_{i:j} \right)
\end{equation}

where \( X_i \) and \( X_{i:j} \) are the design matrices for the main effects and interactions, respectively, and \( \theta_i \), \( \theta_{i:j} \) are their corresponding coefficients.

This logistic model maintains the structure of interactions but within the context of a binary outcome. The optimization process involves minimizing a loss function \( L \), subject to constraints on the coefficients \( \theta \). The group-lasso, a key component of Glinternet, is used to select relevant groups of variables and is defined as:

\begin{equation}
\hspace{-2.5cm} 
\min_{\mu, \beta} \frac{1}{2} \left\| Y - \mu \cdot 1 - \sum_{j=1}^{p} X_j \beta_j \right\|^2_2 + \lambda \sum_{j=1}^{p} \gamma_j \left\| \beta_j \right\|_2
\end{equation}

This formula imposes regularization on the groups of variables, promoting sparsity at the group level.

Glinternet thus provides a robust and flexible framework for modeling interactions in high-dimensional data, ensuring interpretability through the enforcement of strong hierarchical relationships among the variables.

\subsection*{Pretrained lasso}

In Generalized Linear Models (GLMs) and $\ell_1$-regularized GLMs, an offset—a predetermined $n$-vector—can be added as an extra column in the feature matrix with a fixed weight $\beta_j$ of 1 and allows to control for a variable without estimating its parameter \cite{hastie2017generalized}. Additionally, the standard $\ell_1$ norm can be extended to a weighted norm by assigning a penalty factor ${\rm pf}_j \ge 0$ to each feature, modulating the regularization applied \cite{wipf2010iterative}. A penalty factor of zero indicates no penalty, ensuring feature inclusion, while infinity results in feature exclusion, enhancing model flexibility and specificity.

A pretrained lasso model uses both an offset and penalty factor and is fitted in two stages: first, by identifying common features, and then by isolating class-specific features (see Algorithm 1).

\begin{algorithm}
\caption{Pretrained lasso algorithm for ancestry-diverse groups}
\label{alg:PretrainedLassoFixedInputGroups}
Fit an overall LASSO model to the training data with all selected ancestries, selecting $\hat\beta_0$ along the $\lambda$ path that minimizes CV error.
Set a fixed $\alpha \in [0, 1]$. Compute the offset and penalty factor for each input group as follows:
\begin{itemize}
    \item Calculate the linear predictor $X_k\hat\beta_0 + \hat\mu_0$ for each group $k$, and define the offset as $(1-\alpha)\cdot (X_k\hat\beta_0 + \hat\mu_0)$.
    \item Identify the support set $S$ of $\hat\beta_0$. For each feature $j$, define the penalty factor ${\tt pf}_j$ as $(1-\alpha)\cdot [I(j \notin S)\cdot \frac{1}{\alpha} + I(j \in S)]$.
\end{itemize}
For each ancestry $k$, fit an ancestry-specific model.
\begin{itemize}
    \item Using the defined offset and penalty factor, train an L1-penalized logistic regression model on the data of the specific ancestry.
    \item Use these models for group-specific predictions.
\end{itemize}
\end{algorithm}

\subsection*{Study population and diseases}

The UK Biobank is a comprehensive cohort study that gathers data from various locations across the UK. To address the potential variability caused by population structure within our dataset, we limited our analysis to individuals who are not related, adhering to four specific criteria outlined in the UK Biobank's sample quality control file, ukb\_sqc\_v2.txt. These criteria are: (1) inclusion in principal component analysis (as indicated in the used\_in\_pca\_calculation column); (2) individuals not identified as outliers based on heterozygosity and missing data rates (found in the het\_missing\_outliers column); (3) absence of suspected anomalies in sex chromosome number (mentioned in the putative\_sex\_chromosome\_aneuploidy column); and (4) having no more than ten likely third-degree relatives (as per the excess\_relatives column).

We refined our study population by analyzing genetic and self-reported ancestry data, using the UK Biobank's genotype principal components (PCs), self-reported ancestry data (UK Biobank Field ID 21000), and the in\_white\_British\_ancestry\_subset column from the sample QC file. Our focus was on individuals self-identifying as white British (337,129), non-British European (24,905), African (6,497), South Asian (7,831), and East Asian (1,704). The classification into five groups involved a two-stage process, initially using genotype PC loadings to apply specific criteria for each group: (1) white British based on PC1 and PC2 thresholds and inclusion in the white British ancestry subset; (2) non-British European with similar PC thresholds, identified as white but not white British; (3) African, with distinct PC thresholds, excluding identities as Asian, White, Mixed, or Other; (4) South Asian and (5) East Asian, each with unique PC thresholds and exclusions of identifying as Black, White, Mixed, or Other.

We focused on ancestries representing at least 1\% of the dataset, giving us a set of ancestries made of White British, Admixed (Others), Non-British European, South Asian and African.

From this cohort, we extracted data for 96,913 individuals for which we had ancestry and demographic information, metabolites, genotype principal components (PCs), biomarkers and polygenic risk scores (PRS). This gave us an \(n \times p\) data matrix, where \(n = 96,913\) and \(p = 303\). Each row represents an observation, and each column corresponds to a variable.

Our cohort consisted of 80,810 White British (WB), 1,911 South Asians (SA), 1,499 Africans (AF), 6,783 categorized as Admixed (AD) and 5,910 Non-British Europeans(NBE). A detailed breakdown is available in \ref{S3_Fig}. We identified 8 common diseases to use for our study for which we had a minimum of 20 positive cases per ancestry: diabetes (DIA), asthma (AST), chronic renal failure (CRF), osteoarthritis (OST), myocardial infarction (MI), cystistis (CYS), gall stones (GS) and arthritis (ART).

\begin{table}[ht]
\caption{Distribution of cases in the dataset by ancestry and disease.}
\begin{adjustwidth}{-0.2in}{0in}
\begin{tabular}{llrrrrrrrr}
\toprule
Ancestry & Cases & DIA & MI & AST & GS & OST & ART & CYS & CRF  \\
\midrule
WB & Positive & 5479 & 3234 & 10909 & 4103 & 7579 & 4969 & 2725 & 2862  \\
WB & Negative & 75331 & 77576 & 69901 & 76707 & 73231 & 75841 & 78085 & 77948   \\
WB & Total & 80810 & 80810 & 80810 & 80810 & 80810 & 80810 & 80810 & 80810   \\

\midrule
SA & Positive & 460 & 151 & 285 & 69 & 101 & 122 & 56 & 93  \\
SA & Negative & 1451  & 1760 & 1626 & 1842 & 1810 & 1789 & 1855 & 1818   \\
SA & Total & 1911  & 1911 & 1911 & 1911 & 1911 & 1911 & 1911 & 1911   \\
\midrule
AF & Positive & 232  & 30 & 201 & 36 & 86 & 103 & 24 & 83   \\
AF & Negative & 1267  & 1469 & 1298 & 1463 & 1413 & 1396 & 1475 & 1416   \\
AF & Total & 1499  & 1499 & 1499 & 1499 & 1499 & 1499 & 1499 & 1499   \\
\midrule
AD & Positive & 632  & 268 & 932 & 285 & 492 & 379 & 234 & 213   \\
AD & Negative & 6151  & 6515 & 5851 & 6498 & 6291 & 6404 & 6549 & 6570   \\
AD & Total & 6783  & 6783 & 6783 & 6783 & 6783 & 6783 & 6783 & 6783   \\
\midrule
NBE & Positive & 419  & 197 & 768 & 262 & 433 & 279 & 178 & 165   \\
NBE & Negative & 5491  & 5713 & 5142 & 5648 & 5477 & 5631 & 5732 & 5745   \\
NBE & Total & 5910  & 5910 & 5910 & 5910 & 5910 & 5910 & 5910 & 5910   \\
\bottomrule
\end{tabular}
\end{adjustwidth}
\end{table}

\subsection*{Datasets}

The dataset labeled ``All" contains data from all ancestries available in the UK Biobank, including White British (WB) as well as ancestries South Asian, African, Admixed and Non-British European. 
The ``Mix" dataset refers to a combination of White British data with data from one other specific ancestry at a time (e.g. White British and South Asian). This approach allows us to explore the predictive performance of our models when they are trained on data representing the majority population (White British) alongside data from a minority ancestry. 
The term ``ancestry-only data" is used to denote datasets consisting exclusively of data from a single, specific ancestry—such as South Asian (SA) for instance. 
By employing these distinct dataset configurations, our study aims to assess whether training models on all ancestries is beneficial compared to training on White British data and data from a single ancestry.

\subsection*{Training and evaluation}

We started our analysis with a baseline model — an L1-penalized logistic regression — identified as the standard approach for this type of analysis in our literature review \cite{Tibshirani1996}. We used the \textit{glmnet} package \cite{friedman2023glmnet} in R \cite{R} and trained it with the three distinct datasets aforementioned: ancestry-only, Mix, and All (see Figure 3). 

Subsequently, we applied the \textit{glinternet} package in R to train glinternet models \cite{lim2013learning}, selecting age, sex, ancestry, and PRS score as interaction candidates (see Figure 3). Training of glinternet models is conducted on both the Mix and All datasets to assess glinternet's ability to extrapolate patterns across ancestries facing data limitations.

Last, we adopted a two-stage process to train the pretrained lasso model. We used the \textit{ptLasso} package in R \cite{Craig2024} which trains a generic LASSO model using \textit{glmnet}, from which is derived the $\hat{\beta}$ vector. This vector is then used to construct an offset parameter, facilitating the training of ancestry-specific models. A comprehensive pan-ancestry model is subsequently developed by combining these steps (see Figure 3) and is trained like the glinternet models on both the Mix and All datasets.

We divided our dataset into training and testing portions, allocating 80\% for training and 20\% for testing, and employed 3-fold cross-validation to identify optimal hyperparameters for our models. This cross-validation approach, using a limited number of folds, ensures the inclusion of a sufficient number of cases for each ancestry, particularly for those with restricted data (see Figure 3).

We assessed model performance using the test portion of the dataset specific to each ancestry. We calculated the Receiver Operating Characteristic (ROC) curves and ROC-AUC scores \cite{hajian2013receiver} for each model and conducted ROC tests \cite{delong1988comparing} to compare the ROC curves of the glinternet and pretrained lasso models against the baseline models trained on both White British and specific ancestry data. The results include ROC-AUC scores detailed by ancestry and disease, in addition to p-values from the ROC tests and the difference in ROC-AUC ($\Delta$ ROC-AUC) between our models and the corresponding baseline, both trained on the same dataset. All p-values are one-sided unless specified otherwise, as we use a directional hypothesis and expect an improved performance from pretrained lasso and glinternet models compared to our baselines.

\section*{Supporting information}

\paragraph*{S1 Fig.}
\label{S2_Fig}
{\bf Plot of the CV error versus number of main effects and interactions terms along the Glinternet regularization path for osteoarthritis with South Asian individuals.} For the 50 lambda values used for the cross-validation for Glinternet, the number of main effects and interaction terms found by Glinternet is computed, along with the average cross-validation error across the 3 folds. The lowest CV error is at index 35.

\paragraph*{S2 Fig.}
\label{S3_Fig}
{\bf Graph of the interaction network of Glinternet for osteoarthritis amongst South Asian individuals.} The Glinternet model with lowest average CV error across all was chosen. A thicker line indicates a higher interaction coefficient in absolute value. This model was trained on the Mix subset for SA (White British and South Asian data).

\paragraph*{S3 Fig.}
\ref{S3_Fig}
{\bf Diagram of the methodology followed for this project.} 

\paragraph*{S1 Table.}
\label{S2_Table}
{\bf ROC-AUC scores for logistic regression, pretrained lasso and glinternet (African ancestry).} 

\paragraph*{S2 Table.}
\label{S3_Table}
{\bf P-values and $\Delta$ ROC-AUC for pretrained lasso and glinternet against logistic regression (African ancestry.)} 

\paragraph*{S3 Table.}
\label{S4_Table}
{\bf ROC-AUC scores for logistic regression, pretrained lasso and glinternet (South Asian ancestry).} 

\paragraph*{S4 Table.}
\label{S5_Table}
{\bf P-values and $\Delta$ ROC-AUC for pretrained lasso and glinternet against logistic regression (South Asian ancestry).} 

\paragraph*{S5 Table.}
\label{S6_Table}
{\bf ROC-AUC scores for logistic regression, pretrained lasso and glinternet (Admixed ancestry).} 

\paragraph*{S6 Table.}
\label{S7_Table}
{\bf P-values and $\Delta$ ROC-AUC for pretrained lasso and glinternet against logistic regression (Admixed ancestry).} 

\paragraph*{S7 Table.}
\label{S1_Table}
{\bf Distribution of the cases in the dataset by ancestry and disease.} 

\section*{Acknowledgments}

Some of the computing for this project was performed on the Sherlock cluster. We would like to thank Stanford University and the Stanford Research Computing Center for providing computational resources and support that contributed to these research results. The content is solely the responsibility of the authors and does not necessarily represent the official views of the funding agencies; funders had no role in study design, data collection and analysis, decision to publish, or preparation of the manuscript. M.A.R. is in part supported by National Human Genome Research Institute (NHGRI) under award R01HG010140, and by the National Institutes of Mental Health (NIMH) under award R01MH124244 both of the National Institutes of Health (NIH).

\section*{Author Contributions}

\paragraph*{\bf Conceptualization:} Manuel Rivas
\paragraph*{\bf Funding acquisition:} Robert Tibshirani, Trevor Hastie, Manuel Rivas
\paragraph*{\bf Methodology:} Thomas Le Menestrel, Erin Craig, Robert Tibshirani, Trevor Hastie, Manuel Rivas
\paragraph*{\bf Validation:} Thomas Le Menestrel, Erin Craig, Robert Tibshirani, Trevor Hastie
\paragraph*{\bf Writing - original draft:} Thomas Le Menestrel, Manuel Rivas
\paragraph*{\bf Writing - review \& editing:} Thomas Le Menestrel, Erin Craig, Robert Tibshirani, Trevor Hastie, Manuel Rivas

\nolinenumbers

%
%
%

\newpage
\newpage
\bibliography{references}

\end{document}